\documentclass[runningheads]{llncs}
\usepackage[T1]{fontenc}
\usepackage[misc]{ifsym}
\usepackage{graphicx}
\usepackage{natbib} 
\usepackage{amssymb}
\usepackage{amsmath}
\usepackage{bbm}
\usepackage{booktabs}
\usepackage[ruled]{algorithm2e}

\newcommand{\qq}[1]{\text{``}{#1}{\text{"}}}

\begin{document}
\title{Batch Reinforcement Learning from Crowds}
\toctitle{Batch Reinforcement Learning from Crowds}
\author{Guoxi Zhang{\Letter}\inst{1} \and Hisashi Kashima\inst{1,2}}
\toctitle{Guoxi~Zhang\and Hisashi~Kashima}
\institute{Graduate School of Informatics, Kyoto University
\email{guoxi@ml.ist.i.kyoto-u.ac.jp, kashima@i.kyoto-u.ac.jp}\and RIKEN Guardian Robot Project}
\maketitle
\begin{abstract}
A shortcoming of batch reinforcement learning is its requirement for rewards in data, thus not applicable to tasks without reward functions. 
Existing settings for the lack of reward, such as behavioral cloning, rely on optimal demonstrations collected from humans. Unfortunately, extensive expertise is required for ensuring optimality, which hinder the acquisition of large-scale data for complex tasks. 
This paper addresses the lack of reward by learning a reward function from preferences between trajectories. Generating preferences only requires a basic understanding of a task, and it is faster than performing demonstrations. Thus, preferences can be collected at scale from non-expert humans using crowdsourcing. 
This paper tackles a critical challenge that emerged when collecting data from non-expert humans: the noise in preferences. A novel probabilistic model is proposed for modelling the reliability of labels, which utilizes labels collaboratively. Moreover,  the proposed model smooths the estimation with a learned reward function. 
Evaluation on Atari datasets demonstrates the effectiveness of the proposed model, followed by an ablation study to analyze the relative importance of the proposed ideas.
\keywords{Preference-based Reinforcement Learning\and Crowdsourcing}
\end{abstract}
\section{Introduction}

Batch Reinforcement Learning (RL)~\citep{Lange2012} is a setting for RL that addresses its limitation of data acquisition. 
Online RL needs to generate new data during learning, either via simulation or physical interaction. 
However, simulators with high fidelity are not always available, and real-world interactions raise safety and ethical concerns. 
Batch RL instead reuses existing vastly available data, so it has received increasing attention in recent years~\citep{pmlr-v119-pavse20a, pmlr-v119-agarwal20c, DBLP:conf/aaai/GeladaB19,10.5555/3454287.3455342,fujimoto2019benchmarking}.

In batch RL, the data consists of observations, actions, and rewards. 
For example, in recommender systems, the observations are user profiles, and the actions are items to be recommended. 
The rewards are scalars, evaluating actions for their consequences on achieving a given task. 
The mapping from observations and actions to rewards is called a reward function. Often, the sequence of observations, actions, and rewards generated during interaction is called a trajectory. 

While batch RL is a promising data-driven setting, its dependence on rewards can be problematic. 
Many tasks of interest lack a reward function. 
Consider an application to StarCraft, a famous real-time strategy game, for example.
Typically, evaluative feedback is given for the final result of a series of action sequences (i.e., the entire trajectory), not for individual actions about their contributions to the result. 
In the RL literature, the lack of reward has been addressed by inverse RL~\citep{10.1145/1015330.1015430} or Behavioral Cloning (BC) ~\citep{10.5555/2998981.2999127}, which eliminate the need for rewards by leveraging expert demonstrations.  
However, their assumption on demonstrations can be hard to satisfy. 
Optimal demonstrations require extensive expertise. In practice, the competency of human demonstrators differs~\citep{DBLP:conf/iros/MandlekarBSTGZG19},  causing RL and BC algorithms to fail~\citep{mandlekar2021matters}.

This paper addresses the lack of reward signals by learning a reward function from preferences. 
 A \textit{preference} is the outcome of a comparison between two trajectories for the extent the task is solved. 
Compared to demonstrations, providing preferences requires less human expertise. 
For example, demonstrating the moves of professional sports players is difficult, but with general knowledge a sports fan can still appreciate the moves in games. 
Hence, preferences can be collected at scale from a large group of non-expert humans, possibly via crowdsourcing~\citep{10.5555/3122009.3242050}, which is the use of the vast amount of non-expert human knowledge and labor that exists on the Internet for intellectual tasks.
In the RL literature, learning from preferences is discussed as Preference-based RL (PbRL)~\citep{JMLR:v18:16-634}. 
Recent advances show that agents can solve complex tasks using preferences in an online RL setting~\citep{NIPS2017_d5e2c0ad,10.5555/3327757.3327897}. 
This paper extends PbRL to a batch RL setting and focuses on the following challenge: How to learn a reward function from noisy preferences? 

Denoising becomes a major requirement when preferences are collected using crowdsourcing. Crowd workers can make mistakes due to the lack of ability or motivation, thus data generated with crowdsourcing is often very noisy. Similar observation is made when collecting demonstrations from the crowd. ~\citet{pmlr-v87-mandlekar18a,DBLP:conf/iros/MandlekarBSTGZG19} discovers that collected demonstrations hardly facilitate policy learning due to noise in demonstrations. 
However, this challenge has been overlooked by the PbRL community. 
The reason is that, in an online setting, preferences are often collected from recruited collaborators, so their quality is assured. Meanwhile, the present study assumes preferences are collected from non experts, and little is known or can be assured about the annotators. 

\begin{figure}[t]
    \centering
    \includegraphics[width=0.7\linewidth]{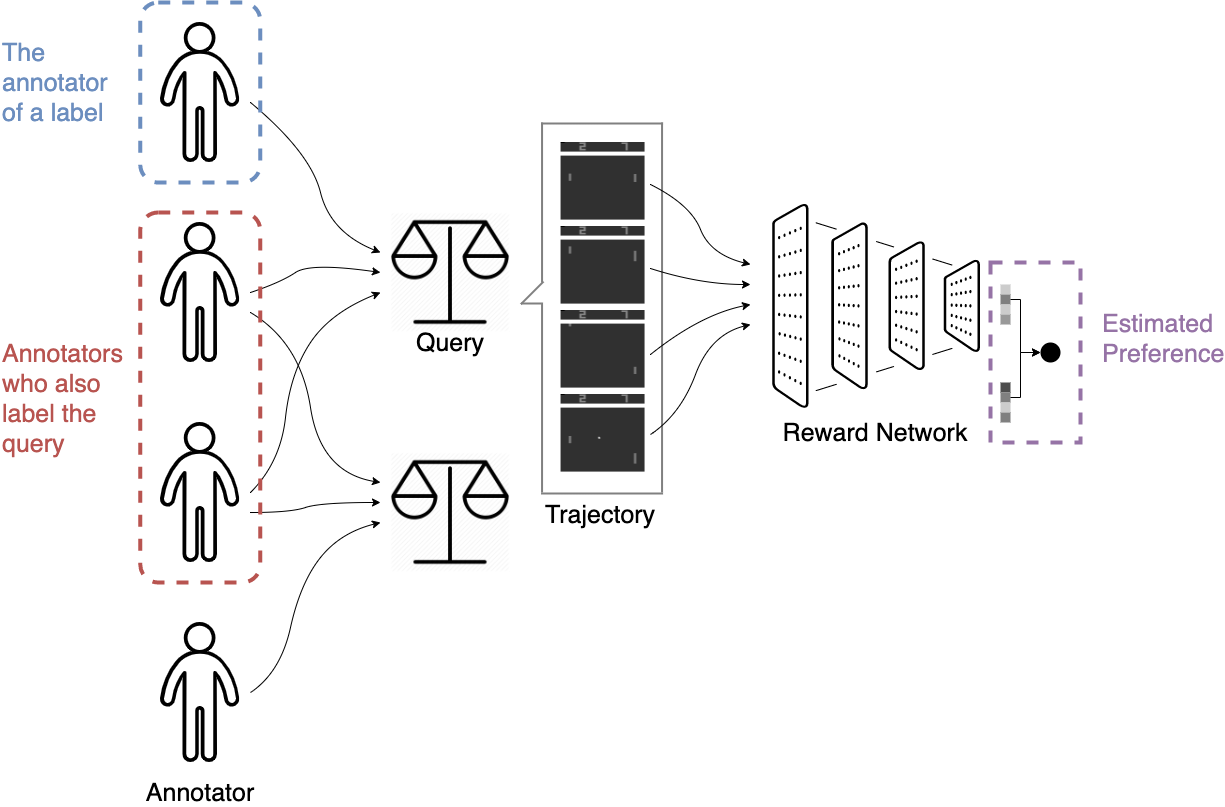}
    \caption{This diagram illustrates information utilized by the proposed DCBT model. An arrow pointing to a query means that the annotator gives a label for the query. When determining the reliability of a label, the DCBT model utilizes: (a) the ID of the annotator of the label (the blue box), (b) the IDs of annotators who also label the query and the labels they give (the red box), and (c) an estimate for preference computed using the learned reward function. Part (b) is our idea of utilizing other labels for the same query to assist in modeling label reliability. Meanwhile, an estimate for preference can be computed with (c), which is also helpful in determining label reliability. }
    \label{fig:DCBT_diagram}
\end{figure}

This paper proposes a probabilistic model, named Deep Crowd-BT (DCBT), for learning a reward function from noisy preferences. 
DCBT assumes that each preference label is potentially unreliable, and it estimates the label reliability from data.
As shown in Figure~\ref{fig:DCBT_diagram}, the idea behind DCBT is to model the correlation of the reliability of a label with its annotator, other labels for the same query, and the estimated reward function. 
The conditional dependency on the annotator models the fact that unreliable annotators tend to give unreliable labels.
Meanwhile, as it is a common practice to solicit multiple labels for the same query, the proposed model collaboratively utilizes labels from different annotators for the same query. 
Yet in practice each query can only be labelled by a small group of annotators given a fixed labelling budget, so DCBT also utilizes the estimated reward function effectively smooths the label reliability. 

A set of experiments on large scale offline datasets~\citep{pmlr-v119-agarwal20c} for Atari 2600 games verifies the effectiveness of DCBT. 
The results show that DCBT facilitates fast convergence of policy learning and outperforms reward learning algorithms that ignore the noise. 
Furthermore, an ablation study is also performed to analyze the relative importance of collaboration and smoothing. 
The contributions of this paper are summarized as follows:
\begin{itemize}
    \item This paper addresses the lack of reward in batch RL setting by learning a reward function from noisy preferences. 
    \item A probabilistic model is proposed to handle the noise in preferences, which collaboratively models the reliability of labels and smooths it with the estimated reward function.
    \item Experiments on Atari games, accompanied by an ablation study, verify the efficacy of the proposed model.
\end{itemize}

\section{Related Work}
Batch RL is a sub-field of RL that learns to solve tasks using pre-collected data instead of online interaction. Efforts have been dedicated to issues that emerge when learning offline, such as the covariate shift~\citep{DBLP:conf/aaai/GeladaB19} and the overestimation of Q function~\citep{10.5555/3454287.3455342}. This paper addresses the lack of reward in batch RL setting, which complements policy learning algorithms.

In the literature of RL, the lack of reward is canonically addressed by either inverse RL~\citep{10.1145/1015330.1015430} or BC ~\citep{10.5555/2998981.2999127}. Inverse RL does not apply as it requires online interactions. BC suffers from the inefficiency of data acquisition and the imperfectness of collected demonstrations. While data acquisition can be scaled up using crowdsourced platforms such as the RoboTurk~\citep{pmlr-v87-mandlekar18a,DBLP:conf/iros/MandlekarBSTGZG19}, the imperfectness of demonstrations remains an issue for BC~\citep{mandlekar2021matters}. Recent results show that BC can work on mixtures of optimal and imperfect demonstrations via collecting confidence scores for trajectories~\citep{pmlr-v97-wu19a} or learning an ensemble model for behavioral policies~\citep{sasaki2021behavioral}. These methods still require large amount of optimal trajectories.

Instead, this paper addresses the lack of reward by collecting preferences from humans. As preferences require less expertise than optimal demonstrations, they can be collected using methods such as crowdsourcing~\citep{10.5555/3122009.3242050}. In the literature of RL, PbRL is shown to be successful for complex discrete and continuous control tasks~\citep{NIPS2017_d5e2c0ad,10.5555/3327757.3327897}. Interested readers may refer to the detailed survey from~\citet{JMLR:v18:16-634}. However, the existing work on PbRL is restricted to clean data in the online RL setting, while crowdsourced data are often noisy~\citep{10.14778/3055540.3055547, AAAI1816102}. While being detrimental to reward learning~\citep{10.5555/3327757.3327897}, noise in preferences remains a overlooked issue. This paper extends PbRL to a batch setting and overcomes the issue of noise in preferences. It extends the propabilistic model proposed by~\citet{10.1145/2433396.2433420} to effectively model label reliability while learning a reward functions from preferences.

\section{Problem Setting}
\subsection{Markov Decision Process}
A sequential decision-making problem is modeled as interactions between an agent and an environment, described as a discounted infinite-horizon MDP: $\langle\mathcal{S}, \mathcal{A}, R, P, \gamma \rangle$, where $\mathcal{S}$ refers to the state space (we interchangeably use \textit{states} and \textit{observations} in this paper), $\mathcal{A}$ is a finite and discrete set of actions, and $R(s,a)$ is the reward function. $P(s'|s, a)$ is the transition probability that characterizes how states transits depending on the action, which is not revealed to the agent. $\gamma\in\mathbb{R}$ is the discount factor.  

Interactions roll out in discrete time steps. At step $t$, the agent observes $s_t\in\mathcal{S}$ and selects action $a_t$ according to a policy $\pi:\mathcal{S}\rightarrow \mathcal{A}$. 
Based on $(s_t, a_t)$, the environment decides next state $s_{t+1}$ according to $P(s_{t+1}|s_t, a_t)$, and the reward $r_t\in \mathbb{R}$ is determined according to $R(s,a)$. 
The objective of policy learning is to find a policy $\pi$ that maximizes $\sum_{t=1}^\infty \gamma^{t-1}R(s_t, a_t)$, the discounted sum of the rewards.

\subsection{Reward Learning Problem}
In this paper, trajectories are assumed to be missing from trajectories. A trajectory can be written as $\eta=(s_1, a_1, \dots, s_{T_c}, a_{T_c})$, where $T_c$ is the length of this trajectory. A learning agent is provided with a set of trajectories and preferences over these trajectories generated by a group of annotators. The $i^{\mathrm{th}}$ sample can be written as a four tuple: $(\eta_{i,1}, \eta_{i,2}, y_i, w_i)$.
The pair $(\eta_{i,1}, \eta_{i,2})$ is the preference query, and $y_i$ is the preference label. $y_{i}=\qq{\succ}$ if in $\eta_{i,1}$ the task is solved better than in $\eta_{i,2}$,
$y_{i}=\qq{\approx}$ if the two clips are equally good, 
and $y_{i}=\qq{\prec}$ if in $\eta_{i,1}$ the task is solved worse than in $\eta_{i,2}$. 
$w_i$ is the ID of the annotator who gave $y_i$. 
Let $N$ be the number of preferences, and let $M$ be the number of annotators. 

The preferences are assumed to be noisy, as the annotators may lack of expertise of commitment. Yet no information other than annotator IDs is revealed to the learning agent. From the preferences, the learning agent aims at learning a reward function $\hat{R}$. The learning problem is summarized as follows:
 
\begin{itemize}
    \item Input: A set of noisy preferences $D = \{(\eta_{i,1}, \eta_{i,2}, y_{i}, w_i)\}_{i=1}^N$, where $\eta_{i,1}$ and $\eta_{i,2}$ are two trajectories, $w_i$ is the identity of the annotator, and $y_i\in\{\qq{\succ}, \qq{\approx}, \qq{\prec}\}$ is the preference label.
    \item Output: An estimated reward function $R: \mathcal{S} \times \mathcal{A}\rightarrow \mathbb{R}$.
\end{itemize}

After learning $R$, the trajectories are augmented with estimated rewards given by $R$. 
They are now in the standard format for off-policy policy learning, and in principle any algorithm of the kind is applicable. 

\section{Proposed Method}
\begin{figure}[t]
    \centering
    \includegraphics[width=0.8\columnwidth]{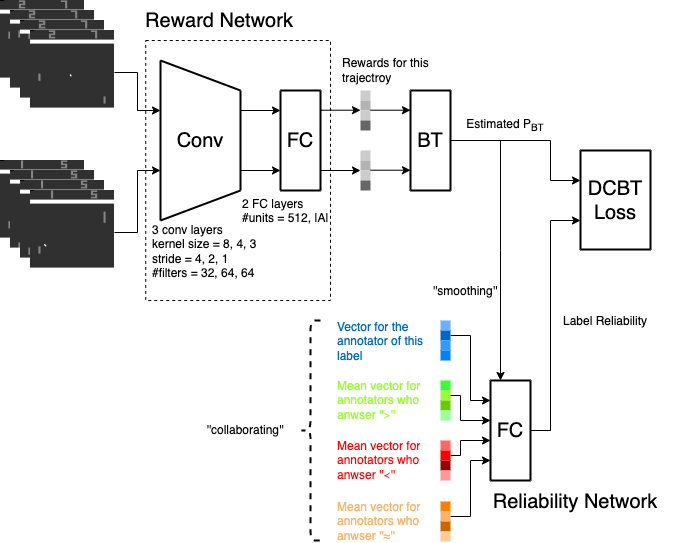}
    \caption{This figure illustrates the architecture for learning reward function with the proposed DCBT model. After learning, the reward function (shown in dashed box) can be used to infer rewards using states and actions of trajectories.}
    \label{fig:network}
\end{figure}
This section presents the proposed Deep Crowd-BT (DCBT) model. A diagram for DCBT is shown in Figure~\ref{fig:DCBT_diagram}, and a pseudocode for reward learning with DCBT is provided in Algorithm~\ref{algo}.

\subsection{Modeling Preferences}
Given a sample $(\eta_{i,1}, \eta_{i,2}, y_{i}, w_i)$, DCBT first computes the rewards of $\eta_{i,1}$ and $\eta_{i,2}$. For each state-action pair $(s,a)$ in $\eta_{i,1}$ and $\eta_{i,2}$, the reward network outputs a scalar $R(s,a)$. This is done by the reward network shown in the upper part of Figure~\ref{fig:DCBT_diagram}. Let $\theta_R$ be the parameters of $R$. For image input, $R$ is parameterized with convolutional neural networks followed by feedforward networks. 

Then the probability of the event ``$\eta_{i,1}\succ\eta_{i,2}$'' is modeled with the Bradly-Terry model (BT)~\citep{10.2307/2334029}:

\begin{equation}
\begin{split}
    \mathrm{P_{BT}}\left(\eta_{i,1} \succ \eta_{i,2}\right) &=\frac{\exp \left(  G(\eta_{i,1})  \right)}{\exp \left(G(\eta_{i,1}) ) \right) + \exp \left(G(\eta_{i,2})  \right)}, \\
    G(\eta_{i,1}) &= \frac{1}{T_c}\sum\limits_{(s,a)\in\eta_{i,1}}R(s,a),\\
    G(\eta_{i,2}) &= \frac{1}{T_c}\sum\limits_{(s,a)\in\eta_{i,2}}R(s,a).\\
\end{split}
\label{bt}
\end{equation}
\noindent The larger $G(\eta_{i,1})$ is, the larger $\mathrm{P_{BT}}(\eta_{i,1} \succ \eta_{i,2})$ becomes. 
$\mathrm{P_{BT}}(\eta_{i,1}\succ\eta_{i,2}) > 0.5$ if $G(\eta_{i,1}) > G(\eta_{i,2})$, that is, the trajectory with a larger sum of rewards is preferred. 
If the preferences are consistent with the task of interest, then maximizing the loglikelihood of $\mathrm{P_{BT}}\left(\eta_{i,1} \succ \eta_{i,2}\right)$ will generate reward function that facilitate learning of competitive policies~\citep{NIPS2017_d5e2c0ad, 10.5555/3327757.3327897}.

\subsection{Handling Noise in Preferences}
When the preferences are noisy, preference labels are not entirely consistent with the underlying task. 
\citet{10.5555/3327757.3327897} point out that policies degenerate for noisy preference data. In literature, the Crowd-BT model~\citep{10.1145/2433396.2433420} is a probabilistic model for modeling noisy pairwise comparisons. It assumes each annotator makes errors with an annotator-specific probability. Denote by $\alpha_w = \mathrm{P}(\eta_{i,1} \succ_w \eta_{i,2} \mid \eta_{i,1} \succ \eta_{i,2})$ the probability that annotator $w$ gives $\qq{\succ}$ for $(\eta_{i,1}, \eta_{i,2})$, when the groundtruth is $\eta_{i,1} \succ \eta_{i,2}$. 
With the Crowd-BT model,
\begin{equation}
\mathrm{P_{Crowd-BT}}(y_i = \qq{\succ}) = \alpha_{w_i}\mathrm{P_{BT}}(\eta_{i,1} \succ \eta_{i,2})+ (1-\alpha_{w_i}) (1 - \mathrm{P_{BT}}(\eta_{i,1} \succ \eta_{i,2})).
\label{crowd-bt}
\end{equation}

In other words, the Crowd-BT model assumes that the reliability of labels from the same annotator is the same and fixed, in regardless of queries being compared. This assumption, however, is inadequate in our case and suffers from at least two reasons. In our case, each query is labeled by multiple annotators. The other labels for the same query and the credibility of their annotators are informative when modeling the reliability of a label, but they are not utilized in the Crowd-BT model. Meanwhile, in practice rich coverage of possible trajectories are required to ensure the generalization of $R$. Thus, under limited labeling budget, each preference query is labeled by a tiny group of annotators, which incurs high variance in labels.

\begin{algorithm}[t]
    \caption{Reward Learning with DCBT}
    \KwIn{$D$, a noisy preference dataset\newline
    $T_\mathrm{INIT}$, the number of graident steps for the initialization phase\newline
    $T_\mathrm{TOTAL}$, the total number of gradient steps\newline
    $T_\mathrm{ALT}$, the period of alternative optimization\newline
    $\beta T_\mathrm{ALT}$ is the number of steps to train $\theta_R$}
    Initialize $\theta_R$, $\theta_W$ and $\theta_\alpha$ randomly \\
    \For{$t=1,2,\dots,T_\mathrm{TOTAL}$}{
        Sample a batch of preference data \\
        \If{$t\leq T_\mathrm{INIT}$}{
            update $\theta_R, \theta_W, \theta_\alpha$ by minimizing $L_{\mathrm{DCBT-INIT}} + \lambda_1 L_\mathrm{IDF} + \lambda_2 L_\mathrm{\ell_1, \ell_2}$
        }\Else{
            \If{$t\ \mathrm{mod}\ T_\mathrm{ALT} < \beta T_\mathrm{ALT}$}{
                update $\theta_R$ by minimizing Equation~\ref{final_obj}.
            } \Else{
                update $\theta_W, \theta_\alpha$ by minimizing Equation~\ref{final_obj}.
            }
        }
    }
    \label{algo}
\end{algorithm}

The proposed DCBT extends Crowd-BT by explicitly overcoming the above-mentioned issues. Instead of per-annotator reliability parameter, it learns a per-sample reliability defined as:
\begin{equation}
\alpha_i=\mathrm{P}(y_i = \qq{\succ} \mid w_i, C_i, \mathrm{P}_{\mathrm{BT}}(\eta_{i,1} \succ \eta_{i,2})).
\label{eq: error_prob}
\end{equation}
\noindent The probability of  $y_i=\qq{\succ}$ can be expressed as
\begin{equation}
\begin{split}
    \mathrm{P_{DCBT}}(y_i = \qq{\succ}) &= \alpha_{i}\mathrm{P_{BT}}(\eta_{i,1} \succ \eta_{i,2})+(1-\alpha_{i})(1-\mathrm{P_{BT}}(\eta_{i,1} \succ \eta_{i,2})).
\end{split}
\label{dcbt}
\end{equation}

For each sample, the reliability network shown in the bottom part of Figure~\ref{fig:DCBT_diagram} outputs label reliability$\alpha_i$. This network is parameterized with the fully-connected layer with sigmoidal activation. Let the parameter of this network be $\theta_\alpha$. This network addresses the drawbacks of the Crowd-BT model by taking as input $w_i$, $C_i$ and $ \mathrm{P}_{\mathrm{BT}}(\eta_{i,1} \succ \eta_{i,2})$.

$C_i$ is a set that contains labels and their annotators for the same preference query $(\eta_{i,1}, \eta_{i,2})$. Specifically,
\begin{equation}
C_i = \{(y_j, w_j) \mid \eta_{j,1} = \eta_{i,1}, \eta_{j,2} = \eta_{i,2}, j\neq i, (\eta_{j,1}, \eta_{j,2}, y_j, w_j)\in D \}.
\end{equation}
\noindent Using $C_i$, the reliability network utilizes crowdsourced preferences \textit{collaboratively}. The intuition is that, for the same query, labels from other annotators and the credibility of these annotators provide useful information for modeling $\alpha_i$. A label might be reliable if it is consistent with other labels, especially with those from credible annotators. Moreover, the label values matter. For example, suppose $(\qq{\prec}, w_j)\in C_i$, and $w_j$ is a a credible annotator. Then while both $\qq{\succ}$ and $\qq{\approx}$ are inconsistent with $\qq{\prec}$, the latter should be a more reliable one. The reliability network relies on a worker embedding matrix $\theta_W\subset \mathbb{R}^d$. It takes as input the embedding vector of $w_i$. For $C_i$, it groups the label-annotator pairs into three groups by labels values. Then it computes the mean vector of worker embedding vectors in each group and concatenates these mean vectors together. For example, the red vector in the bottom of Figure~\ref{fig:DCBT_diagram} corresponds to the mean vector of annotators who answer $\qq{\prec}$. Zero vector is used when a group does not contain any annotators.

Meanwhile, the reliability network utilizes $\mathrm{P_{BT}}(\eta_1 \succ \eta_2)$ to address the variance in labels. The present study claims that label reliability also depends on the difficulty of queries. A label $y_i=\qq{\approx}$ is less reliable if $\eta_1$ is significantly better than $\eta_2$, when compared to the case in which $\eta_1$ is only slightly better than $\eta_2$. Thus the reliability network utilizes $\mathrm{P_{BT}}(\eta_1 \succ \eta_2)$ in determining the reliability of $y_i$. As the $R$ summarizes information from all annotators, utilizing $\mathrm{P_{BT}}(\eta_1 \succ \eta_2)$ effectively \textit{smooths} the labels collected for each queries.

\subsection{Learning}
The parameter $\theta_R$, $\theta_W$, and $\theta_\alpha$ can be learned by minimizing the following objective function:
\begin{equation}
    \begin{split}
         L_{\mathrm{DCBT}}(\theta_R, \theta_W, \theta_\alpha) = -\frac{1}{N}\sum_{i=1}^N \left[\tilde{y_i} \log(\mathrm{P_{DCBT}})  + (1-\tilde{y_i})\log(1-\mathrm{P_{DCBT}})\right],\\
    \end{split}
\end{equation}
\noindent where $\mathrm{P_{DCBT}}$ is a short-hand notation for $\mathrm{P_{DCBT}}(y_i = \qq{\succ})$. $\tilde{y_i}$ equals to 1 if $y_i=\qq{\succ}$, 0.5 if $y_i=\qq{\approx}$ and 0 if $y_i=\qq{\prec}$. Note that $\mathrm{P_{DCBT}}$ is a function of $\theta_R, \theta_W$ and $\theta_\alpha$ for a fixed dataset $D$. This dependency is omitted in notations for simplicity.

As mentioned by~\citet{10.1145/2433396.2433420}, there is an identifiability issue for learning from pairwise comparisons. For a query $(\eta_{i,1}, \eta_{i,2})$, $\mathrm{P_{BT}}(\eta_{i,1} \succ \eta_{i,2})$ does not change when adding an arbitrary constant $C\in\mathbb{R}$ to $G(\eta_{i,1})$ and $G(\eta_{i,2})$. Preliminary experiments show that reward network tends to output large values, a similar situation with the over-fitting problem in supervised learning. Following~\citet{10.1145/2433396.2433420}, our learning algorithm utilizes a regularization term $L_{\mathrm{reg}}$ to restrict reward values around zero. Moreover, $\ell_1$ and $\ell_2$ regularization are also helpful to reduce over-fitting.
\begin{equation}
    \begin{split}
        &L_{\mathrm{reg}}(\theta_R)=  \\
        &- \frac{1}{2N}\sum_{i=1}^N\sum_{k=1}^2  \left[\log \left( \frac{\exp \left(G(\eta_{i,k})\right)}{\exp \left(G(\eta_{i,k})\right) + 1} \right)  + \log \left( \frac{1}{\exp \left(G(\eta_{i,k})\right) + 1}\right)\right].
    \end{split}
    \label{reg}
\end{equation}

The overall objective function can be compactly written as
\begin{equation}
    L(\theta_R, \theta_W, \theta_\alpha) = L_\mathrm{DCBT} + \lambda_1 L_\mathrm{reg} + \lambda_2 L_\mathrm{\ell_1, \ell_2},
    \label{final_obj}
\end{equation}

\begin{figure*}[t]
    \centering
    \includegraphics[width=\textwidth]{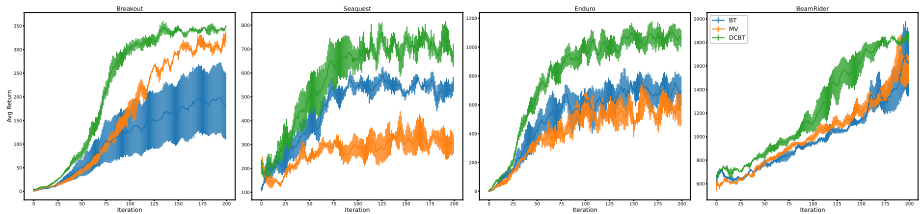}
    \caption{Performance comparison among the proposed DCBT model, BT model, and MV. The curves are for the average returns obtained during training a QR-DQN agent. The QR-DQN agents trained with the proposed method performs better than those trained with BT and MV for game \textit{Seaquest} and \textit{Enduro}. Moreover, for all of the games, the proposed DCBT model enables faster policy learning convergence, which is also desirable in practice.}
    \label{fig:avg_returns}
\end{figure*}

Besides, an initialization phase is required before optimizing Equation~\ref{final_obj}. This phases initializes $R$ by maximizing the loglikelihood of the BT model. The intuition is that by regarding all labels as correct, the reward network can attain intermediate ability in modeling preferences. It also initializes $\theta_\alpha$ and $\theta_W$ by minimizing the cross entropy between $\alpha_i$ and a Bernoulli distribution with parameter $\bar\alpha$. This follows the initialization procedure for the Crowd-BT model. Formally, the objective function for initialization phase, $L_\mathrm{DCBT-INIT}$, is defined as:
\begin{equation}
    \begin{split}
    &  L_{\mathrm{DCBT-INIT}}(\theta_R, \theta_W, \theta_\alpha) =\\
    &-\frac{1}{N}\sum_i^N[\tilde{y_i}\log \mathrm{P_{BT}}+ (1-\tilde{y_i})\log(1-\mathrm{P_{BT}})+ \bar{\alpha}\log(\alpha_i) + (1-\bar{\alpha})\log(1-\alpha_i)],
    \end{split}
\end{equation}
\noindent where $\mathrm{P_{BT}}$ is a short-hand notation for $\mathrm{P_{BT}}(y_i=\qq{\succ})$. Note that it is a function of $\theta_R$, and $\alpha_i$ is a function of $\theta_R, \theta_W$ and $\theta_\alpha$. $\bar{\alpha}$ is a hyper-parameter, which is set to $0.99$. The gradients from the latter two terms of $L_{\mathrm{DCBT-INIT}}$ to $\theta_R$ are blocked, as they use pseudo labels instead of true labels.

After initialization, the objective function in Equation~\ref{final_obj} is minimized with an alternative scheme, similar to the method described by \citet{10.1145/2433396.2433420}. At each round, $\theta_R$ is first optimized for several gradient steps while keeping $\theta_W$ and $\theta_\alpha$ fixed. 
Then, $\theta_W$ and $\theta_\alpha$ are optimized for several steps while $\theta_R$ is fixed. 
The entire algorithm is given in Algorithm~\ref{algo}.

\section{Evaluation}% with Synthetic Preference}
\subsection{Setup}
The present study utilizes the benchmark datasets published by \citet{pmlr-v119-agarwal20c} for experimental evaluations; they contain trajectories for Atari 2600 games collected during training a DQN agent. Due to limited computation resources, four of the games used in existing work for online PbRL~\citep{10.5555/3327757.3327897} are selected. In practice, trajectories are too long to be processed, so they are truncated to clips of length 30 ($T_c=30$). From each game 50,000 clips of trajectories are randomly sampled.
Queries are sampled randomly from these clips.

To have precise control for error rates, preferences are generated by 2,500 simulated annotators. Each annotator generates a correct preference label with a fixed probability sampled from $\mathrm{Beta}(7,3)$. When making mistakes, an annotator selects one of the two incorrect labels uniformly at random. Each annotator labels at most 20 queries, which means the number of preferences does not exceed 50, 000. 
In our experiments, each query is annotated by at most ten different annotators. 

Algorithms for reward learning are evaluated using the performance of the same policy policy-learning algorithm. Reward functions are learned using different reward learning algorithms on the same set of noisy preferences. 
Then, the reward functions are utilized to compute rewards for learning policies using the same policy-learning algorithm, which means that the performance of the policies reflects the performance of the reward-learning algorithms. 
For the policy learning algorithm, the quantile-regression DQN algorithm (QR-DQN)~\citep{Dabney2018} is adopted due to its superior performance. 
A recent empirical analysis shows that this algorithm yields the state-of-the-art performance in batch RL settings~\citep{pmlr-v119-agarwal20c}. 
Obtained policies are evaluated in terms of the average return obtained per episode. 
Experiments are repeated three times on three different sets of trajectories of a game. 
The mean values of returns and their standard error are reported.

\subsection{Implementation Details}
Figure~\ref{fig:network} shows the network structure used for reward learning using the DCBT model. 
The other reward learning algorithms utilize the same reward network. 
The convolutional network part of this architecture is the same as that of the QR-DQN agent released by \citep{pmlr-v119-agarwal20c}.

The QR-DQN agent is trained for 200 training iteration. 
In each training iteration, to speed up training, the agent is continuously trained for 62,500 \textbf{gradient steps}. 
In its original implementation~\citep{pmlr-v119-agarwal20c}, agents are trained for 250,000 \textbf{environment steps}, during which parameters are updated every four environment steps. 
Therefore, in our evaluations for each iteration, agents are trained for the same number of gradient steps as \citep{pmlr-v119-agarwal20c} performed. 
Except for this difference, the other hyper-parameters are not altered.

\begin{figure*}[t]
    \centering
        \includegraphics[width=\textwidth]{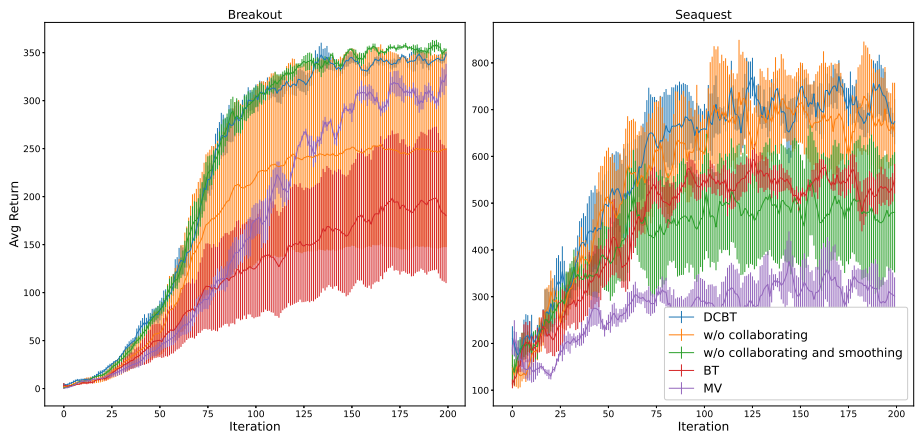}
        \caption{The results of the ablation study for DCBT. For \textit{Breakout}, removing annotator collaborating decreases performance, but further removing smoothing improves performance. For \textit{Seaquest}, removing collaborating results in little difference, but further removing smoothing decreases performance. These results show that combining the two ideas can effectively handle noise in preferences and overcome drawbacks of the two ideas.}
    \label{fig:ablation}
\end{figure*}

\subsection{Alternative Methods}
For all of the four games, the proposed DCBT model is compared with the following two baseline methods.

\paragraph{BT Model.} This method regards all preferences as correct ones, and utilizes the BT model to learn $\hat{R}$. It is the method used in the recent work for online PbRL setting~\citep{NIPS2017_d5e2c0ad,10.5555/3327757.3327897}.

\paragraph{Majority Voting (MV).} This method counts the occurrence of different labels for the same query. The label with the maximum count is chosen as the estimated label. 
Ties are broken randomly. 
Using the estimated labels, $\hat{R}$ is learned with the BT model. 
The estimated labels generated by this method still contain noise, but they are less noisy than the original labels used by the BT method.

\vskip\baselineskip
An ablation study is carried out to analyze the effect of annotator collaborating and smoothing in modeling label reliability. For game \textit{Breakout} and \textit{Seaquest}, the DCBT model is compared with the following two methods.

\paragraph{w/o collaborating.} 
This is a variant of the proposed method that ignores other annotators who also label the same query.

\paragraph{w/o collaborating and smoothing.} 
Not only \textit{w/o collaborating}, the estimated rewards are also ignored. 
Note that this method only considers the identity of the annotator of a query, so it is equivalent to the Crowd-BT model.

\subsection{Results}
Figure~\ref{fig:avg_returns} shows the results for all the four games. 
For the two games, \textit{Seaquest} and \textit{Enduro}, the proposed DCBT achieves the best final performance, which means that DCBT successfully generates reward functions that align with the tasks of interest. 
Meanwhile, for the other two games, \textit{Breakout} and \textit{BeamRider}, using MV results in slower convergence but close final performance. 
Thus, for these two games, the MV method can generate reward functions aligned with the tasks of interest, but such reward functions hinder fast convergence. 
Only for \textit{BeamRider}, the BT method has a similar final performance as the  DCBT model. 
In addition to improved final performance, faster convergence is also a desirable property in practice, especially in scenarios with limited resources. 
From this perspective, the proposed DCBT model also outperforms its alternatives.

Figure~\ref{fig:ablation} shows the ablation study results for DCBT on \textit{Breakout} and \textit{Seaquest}. 
For \textit{Breakout}, removing the annotator collaborating (shown in orange) decreases its performance, although it is still better than the BT method (shown in red). Hence for this game, label collaborating plays an important roll for DCBT.
Interestingly, further removing the smoothing (shown in green) boosts the performance, which is even slightly better than the DCBT model. 
Since this is a rather simple model compared to DCBT, the increase in performance might be due to less overfitting. 

For \textit{Seaquest}, removing annotator collaborating hardly affects the performance. 
Furthermore, removing smoothing significantly decreases the performance. 
So for this game, the idea of smoothing might be important for the performance of the proposed DCBT model.

Our experimental results show that, the efficacy of annotator collaborating and smoothing might be task specific, but their drawbacks can be overcome by combining them together. 
Further investigation of the reasons is left as future work.

\section{Conclusion}
This paper address the lack of reward function in batch RL setting. Existing settings for this problem rely on optimal demonstrations provided by humans, which is unrealistic for complex tasks. So even though data acquisition is scaled up with crowdsourcing, effective policy learning is still challenging. This paper tackles this problem by learning reward functions from noisy preferences. Generating preferences requires less expertise than generating demonstrations. Thus they can be solicited from vastly available non-expert humans. A critical challenge lies in the noise of preferences, which is overlooked in the literature of PbRL. This challenge is addressed with a novel probabilistic model called DCBT. DCBT collaboratively models the correlation between label reliability and annotators. It also utilizes the estimated reward function to compute preference estimates, which effectively smooths labels reliability. Evaluations on Atari 2600 games show the efficacy of the proposed model in learning reward functions from noisy preferences, followed by an ablation study for annotator collaborating and smoothing. Overall, this paper explores a novel methodology for harvesting human knowledge to learn policies in batch RL setting.

Our ablation study indicates the occurrence of over-fitting, which means the reward model might overly fit states and actions in preferences. How to inject induction bias to overcome such an effect is an interesting future work.  Moreover, while there are 50 million states and actions for each game, only 50,000 are covered in queries. Under a fixed labelling budget, how to more effectively generate queries is an important issue.

\section*{Acknowledgment}
This work was partially supported by JST CREST Grant Number JPMJCR21D1.

\bibliography{crowd_pbrl} 
\bibliographystyle{abbrvnat}
\end{document}